# Exact Inference of Hidden Structure from Sample Data in Noisy-OR Networks

**Michael Kearns**
AT&T Labs
180 Park Avenue, Room A235
Florham Park, New Jersey 07932
mkearns@research.att.com

**Yishay Mansour**
Tel Aviv University
Department of Computer Science
Tel Aviv, Israel
mansour@math.tau.ac.il

## 1 Introduction

In the literature on graphical models, there has been increased attention paid to the problems of learning hidden structure (see Heckerman [H96] for a survey) and causal mechanisms from sample data [H96, P88, S93, P95, F98]. In most settings we should expect the former to be difficult, and the latter potentially impossible without experimental intervention. In this work, we examine some restricted settings in which the ideal can be obtained: efficient algorithms that *perfectly* reconstruct the hidden structure solely on the basis of observed sample data.

In our framework, we assume that the unknown "target" network is a two-layer, noisy-OR network meeting a number of assumptions detailed in the paper; briefly, the assumptions limit the "fan-in" of each output node (the number of inputs that can influence that node) and the number of possible values for the weights in the network. Learning algorithms observe independent draws of the output units only; the values of the input units are always unobserved. Rather than just approximate the output distribution (a perfectly reasonable goal), our algorithms *exactly* reconstruct the directed graph from the inputs to the outputs (the hidden structure), and do so in time polynomial in the size of the target network (and other parameters detailed below).

There are two main ideas behind our algorithms:

- The integration of accurate structural information about many "small subnetworks" of the target network in order to obtain the correct structure for the entire network.

- The acquisition of accurate structural information about the small subnetworks from the passively observed values on the outputs.

The strength of our results is indicative of the strength of our assumptions, as we do not expect results of this type to be possible in all but the most fortunate situations. Nevertheless, the assumptions do not trivialize the problem — there are still exponentially many networks meeting our restrictions — and we hope that some of the underlying mathematical tools we introduce may lead to more widely applicable heuristics.

The outline of the paper is as follows. In Section 2, we give definitions for the types of networks we examine, and introduce the restrictions on them that we will require. In Section 3, we introduce an abstraction that we call *Subnetwork Equivalence Queries* that allows us to describe the first main idea behind our algorithms, the integration of accurate structural information about many small pieces of the unknown network. Section 4 examines the conditions required to implement such queries from sample data, and gives the overall description of our algorithms.

## 2 Definitions and Preliminaries

We will use the standard definitions for two-layer noisy-OR belief networks. Such a network has $n$ inputs $X_1, \ldots, X_n$ and $m$ outputs $Y_1, \ldots, Y_m$. For each input $X_i$ and output $Y_j$ there is an associated *weight* $\eta_{ij} \in [0, 1]$, and we say that $X_i$ and $Y_j$ are *connected* if and only if $\eta_{ij} \neq 1$. In our framework, the inputs $X_i$ will be *hidden* variables, whose values are *never* observed in the data available to our learning algorithms. The outputs $Y_j$ will be fully observable.

With each output $Y_j$ we associate the set $S_j$ of (indices of) inputs $X_i$ that are connected to $Y_j$, and we say the network has *fan-in* $k$ if $|S_i| \leq k$ for all $1 \leq i \leq m$.

We use the standard definition for the conditional dis-tribution of $Y_j$ given values for its parents:

$$\mathbf{Pr}[Y_j = 0 | X_1, \ldots, X_n] = \prod_{i=1}^{n} \eta_{ij}^{X_i} = \prod_{i \in S_j} \eta_{ij}^{X_i}.$$

Another way of describing the way such a network generates a distribution on its outputs, given values for the inputs, is as follows. We associate with each input $X_i$ and output $Y_j$ that are connected a binary random variable $R_{ij}$, where $\mathbf{Pr}[R_{ij} = 0] = \eta_{ij}$. The output variable $Y_j$ is set to

$$Y_j = \vee_{i \in S_j} X_i R_{ij}. \tag{1}$$

Thus, we define a *noisy-OR network* $N$ to consist of the connection parameters $\eta_{ij}$ between every input $X_i$ and output $Y_j$. The network $N$ does not yet specify a probability distribution; to specify the joint distribution defined by $N$ it remains only to assign independent *biases* $p_i$ to the input units $X_i$. In this paper, we will restrict our attention to distributions obtained by setting all of the $p_i$ to some common value $p$, and we shall denote the resulting joint distribution on the *outputs* $Y_j$ by $N_p$.

With the networks of interest now specified, we can describe the learning model we will study. Our learning algorithms will be given independent output draws from a distribution defined by an unknown two-layer noisy-OR *target* network $N_p$. Each draw observed by the learning algorithm consists of values for the outputs $Y_1, \ldots, Y_n$ *only* — the values for the inputs $X_1, \ldots, X_m$ that generated the observed outputs are always hidden. In the terminology of the graphical models literature, we are in the *partially observed data, unknown structure* setting.

A perfectly reasonable goal for a learning algorithm in this setting would be to use the data to learn an *approximation* of the unknown target distribution over the $Y_j$ only, since these are the only variables that are observed. In such a model, we do not ask the learning algorithm to explicitly derive the assumed causal relationships between the $X_i$ and the $Y_j$. Indeed, the learning algorithm is free to not represent the $X_i$ at all, since only an accurate approximation of the output distribution is required. There are many works that either implicitly or explicitly take this view of distribution learning. The closest in spirit to the current work is [K94], in which a "PAC-like" model for distribution learning was formalized.

In this paper, however, we study a much more demanding criterion for learning. Under some strong assumptions on the unknown target network generating the observed output draws, we give algorithms that will, with high probability, *exactly recover* the structure of the target network, and furthermore, will do so in time *polynomial* in the size of the target network. Thus, not only will the resulting approximation of the output distribution be *perfect* (Kullback-Leibler divergence 0 to the target distribution), but the true underlying directed graph of the target network will be inferred. This can be viewed as a demonstration of a restricted setting in which it is in fact possible to recover the exact structure on the basis of passively observed data, a task that under general circumstances is difficult or impossible.

As the reader will suspect, in order to obtain such strong results, we will require a number of restrictions on the general two-layer noisy-OR networks we have defined. Our main restrictions will be:

- *Identical* input biases. Thus, $p_i = p \in [0, 1]$ for all $1 \leq i \leq m$. Furthermore, we will eventually see that our algorithms work not for all values of $p$, but only "most" values.

- *Bounded* fan-in. Thus, $|S_j| \leq k$ for all $1 \leq j \leq n$.

- Restrictions on the allowed weights $\eta_{ij}$, discussed below.

The restrictions we require on the $\eta_{ij}$ can be expressed in terms of the number of distinct values the $\eta_{ij}$ are allowed to assume. We say that a class $C$ of networks has $\ell$ *weight values* if there is a finite set $A$ of values in $[0, 1]$ such that $|A| \leq \ell$, and for any network in $C$ and any weight $\eta_{ij}$ in the network, $\eta_{ij} \in A$. We will also study the special case of this restriction in which every value in $A$ is an integer multiple $t\beta$ of some small fixed value $\beta \in [0, 1]$, for some natural number $t$, in which case we say that $C$ has $\ell$ *weight values multiplying* $\beta$. Such a restriction would arise naturally, for example, from discretizing continuous weights into the $\ell$ possible multiples of a resolution parameter $\beta = 1/\ell$.

Now that we have spelled out the various restrictions we will require, let us quickly reiterate some of the remarks made in the introduction about these restrictions. First of all, let us note that despite their strength, these restrictions in no way render the problem of learning the allowed classes of networks trivial from the combinatorial point of view. For example, even with identical input biases, constant fan-in, and only one weight value, the number of $m$-input, $n$-output networks remains exponential in $m$ and $n$ — there are simply exponentially many directed graph *structures* that generate unique distributions. It is precisely this structure that our algorithms will recover *exactly* (with high probability). On the other hand, the very strength of our results — exact inference of the structure in polynomial time — is indicative of restrictions that are unlikely to be met in all but the most limited settings. The results we describe are thus primarily of theoretical interest. Our hope is that some of the mathematical ideas and observations presented will point the way to related heuristics that may be more applicable.

## 3 Subnetwork Equivalence Queries

As mentioned in the introduction, there are two distinct main ideas behind our algorithms:

- The integration of accurate structural information about many "small subnetworks" of the target network in order to obtain the correct structure for the entire network.

- The acquisition of accurate structural information about the small subnetworks from the passively observed values on the outputs.

It turns out that these two ideas can be separated fairly cleanly and fruitfully from a technical viewpoint. In particular, it will be useful to describe our algorithm first in terms of an abstraction we will call *Subnetwork Equivalence Queries (SEQ's)*. This can be thought of as a subroutine or an "oracle" that, given any set of outputs $\mathcal{Y}$ of the target network, and a proposed directed graph between the inputs and the outputs in $\mathcal{Y}$, indicates whether or not the proposed substructure is structurally equivalent to that in the target network. Thus, we propose a candidate for the subnetwork *induced* by $\mathcal{Y}$ (in the graph-theoretic sense), and are told YES if this candidate is correct and NO otherwise.

We will first describe our algorithms assuming we have a subroutine for SEQ's. However, the reason for introducing SEQ's is the hope that we can actually *implement* them solely on the basis of observed data at the outputs of the network. We will later see that this can be achieved *efficiently* only for "small" SEQ's — that is, SEQ's in which the set $\mathcal{Y}$ has small cardinality. Thus, in the remainder of this section, we describe algorithms using small SEQ's that exactly recover the structure of the target network. In Section 4, we tackle the more technical and statistical problem of how to implement small SEQ's under various restrictions on the unknown target noisy-OR network. In any case, algorithms using small SEQ's seems to provide a useful abstraction: for any class of networks for which one can implement SEQ's — whether by sample data, experimentation, or other means — the algorithms of this section are applicable.

### 3.1 Network Equivalence and Basic Blocks

We define two noisy-OR networks to be *structurally equivalent* if they are identical up to renaming of the input variables. More precisely, two noisy-OR networks $N^1$ and $N^2$ over inputs $X_1, \ldots, X_m$ and outputs $Y_1, \ldots, Y_n$ and with weights $\eta_{ij}^1$ and $\eta_{ij}^2$ respectively, are equivalent if there exists a permutation $\pi$ of the inputs such that if $\pi(i) = j$, then $\eta_{il}^1 = \eta_{j\ell}^2$, for any output $\ell$. The goal of our algorithms is to find a network structurally equivalent to the unknown target network.

It is not hard to see that if two networks are structurally equivalent, then for any input bias $p$, they generate identical distributions on their outputs. We will eventually see that the converse is not true (see Figure 1), which will complicate the conditions we require on the unknown network.

Given any noisy-OR network, it will prove useful to partition its inputs into sets that we will call *basic blocks*. Informally, a basic block consists of all those inputs that influence the outputs in an identical manner. Formally, for each input $X_i$ we define the set $T_i$ to contain the (indices of) outputs $Y_j$ such that $i \in S_j$ (that is, $\eta_{ij} \neq 1$). Thus, $T_i$ consists of just those outputs that $X_i$ influences. Then we say that $X_i$ and $X_{i'}$ are in the same *basic block* if and only if $T_i = T_{i'}$, and for every $j \in T_i$, $\eta_{ij} = \eta_{i'j}$. Clearly, the basic blocks define a partition of the input variables.

Now for any basic block $B \subseteq \{X_1, \ldots, X_m\}$ of a noisy-OR network, there are a number of ways of "naming" or specifying $B$. One is obviously by the subset of the $X_i$ in $B$, which allows for the possibility of $2^m$ distinct basic block names. The following simple lemma shows that for limited fan-in networks, there is a much more succinct way of specifying the basic blocks.

**Lemma 3.1** *([K94]) Let $N$ be a noisy-OR network on inputs $X_1, \ldots, X_m$ and outputs $Y_1, \ldots, Y_n$, and let the fan-in of $N$ be bounded by $k$. For any $j$, let $S_j$ be the set of inputs connected to output $Y_j$, as defined above. Then any basic block $B$ of inputs is equal to the intersection of $k$ of the $S_j$ and their complements — that is, there exists $j_1, \ldots, j_k$ such that*

$$B = S_{j_1} \cap \cdots \cap S_{j_\ell} \cap \bar{S}_{j_{\ell+1}} \cap \cdots \cap \bar{S}_{j_k} \qquad (2)$$

To see this, first note that a basic block $B$ is simply a collection of inputs, each of which is connected to exactly the same set of outputs. Without loss of generality, let the outputs that $B$ is connected to be

$Y_1, \ldots, Y_r$. Then we clearly have

$$B = S_1 \cap \cdots \cap S_r \cap \bar{S}_{r+1} \cdots \cap \bar{S}_n. \qquad (3)$$

In particular, $B$ is contained in $S_1$. If $B = S_1$, we are finished. Otherwise, by Equation (3), we must be able to choose one of $S_2, \ldots, S_r, \bar{S}_{r+1}, \ldots, \bar{S}_n$, and by intersecting with $S_1$, reduce the remaining set size by at least 1 while getting "closer" to $B$. But since $S_1$ has only $k$ elements to begin with, after only $k - 1$ intersections of the $n$ given in Equation (3), the resulting intersection will be equal to $B$.

### 3.2 An Incremental Algorithm Using SEQ's

Armed with the notion of basic blocks, we can now describe our algorithms at a high level. First we make the behavior of SEQ's more precise. The input to an SEQ consists of a two-layer noisy-OR network, complete with weights, defined on all the inputs $X_1, \ldots, X_m$ and on just a subset $\mathcal{Y}$ of the outputs $Y_1, \ldots, Y_n$. The SEQ returns YES if the input network is structurally equivalent to the subnetwork induced by the target network on $\mathcal{Y}$. Otherwise, the SEQ returns NO.

For now, we simply assume that we have access to SEQ's, and describe an algorithm for exactly recovering the structure of the unknown target network. As we have already mentioned, however, we will eventually show various conditions under which it is possible to implement SEQ's given only access to samples from the target output distribution. Since the complexity of this implementation will depend crucially on the size of the subnetworks on which the SEQ's are made, we here give an algorithm that only makes SEQ's on small networks.

Let us here reintroduce the assumptions on the target network that we will exploit — namely, that the target network has identical (and known) input biases $p$, fan-in bounded by $k$ and at most $\ell$ weight values, which we also assume are known. (In order to successfully *implement* SEQ's, we will later examine some additional restrictions on the parameters, but these will suffice for now.) Under these conditions, there is a simple incremental algorithm for exactly recovering the target network from SEQ's that we will use as our starting point, and then modify.

The simple incremental algorithm proceeds as follows: assuming for induction that the target network restricted to just the outputs $Y_1, \ldots, Y_{j-1}$ (that is, the network between *all* of the inputs $X_1, \ldots, X_m$ and just these first $j - 1$ outputs) has been perfectly reconstructed, the algorithm proceeds to "add" the structure on $Y_j$. There are at most $n^k$ choices for the set $S_j$ of inputs connected to $Y_j$; for each such choice, there are at most $\ell^k$ choices for the weights on these connections. For each of the resulting $n^k \ell^k$ ways of "wiring" $Y_j$ into the network reconstructed so far, we can then make an SEQ on the entire proposed network on the inputs and $Y_1, \ldots, Y_j$. Clearly one of the queries will return YES, indicating that the structure is correct, and we can proceed to the next output.

This incremental algorithm would make on the order of $n(n^k \ell^k)$ SEQ's, each on a network with as many as $n$ outputs. The size of such queries would result, as we will see in the next section, in a final implementation that required time exponential in $n$ in order to reconstruct the network from sample data. We now describe an improved algorithm that makes SEQ's whose size depends only on $k$.

Suppose we have reconstructed the network through $Y_1, \ldots, Y_{j-1}$, and we divide the $m$ inputs into the basic blocks defined by these $j - 1$ outputs. Clearly, in order to decide how $Y_j$ should be wired, it suffices to know for every basic block $B$ how many inputs are in $B \cap S_j$ — we already know that every input in $B$ is identically connected to the outputs $Y_1, \ldots, Y_{j-1}$, so we simply need to know how many of these are connected to $Y_j$ (and of course, with what weight), and how many are not. Notice that the introduction of $Y_j$ is naturally "breaking" each previous basic block into at most $\ell + 1$ new basic blocks, according to the connectivity to $Y_j$, and the appropriate weight value.

Now the important point is that by Lemma 3.1, if $Y_j$ breaks one of the current basic blocks $B$ into two or more new basic blocks, it must already do so in the subnetwork induced by the $k$ outputs comprising the succinct "name" of $B$. More precisely, in order to determine the connectivity from the basic block $B$ to the output $Y_j$, we can proceed as follows: take the $k$ outputs $Y_{j_1}, \ldots, Y_{j_k}$ from $Y_1, \ldots, Y_{j-1}$ yielding Equation (2) for $B$, and look at the subnetwork induced by these $k$ outputs — this will consist of these $k$ outputs, their inputs (of which there are at most $k^2$), and the connections between them. We now consider all possible ways of adding the new output $Y_j$ to this subnetwork — that is, all possible ways of choosing up to $k$ inputs to $Y_j$ from among the $k^2$ inputs of $Y_{j_1}, \ldots, Y_{j_k}$, with the remaining inputs to $Y_j$ being "new" inputs. For each such choice (of which there are are most $(k^2)^k = k^{2k}$), we will make an SEQ on the resulting subnetwork, and one of them must return YES. For this choice, we can see how many inputs in $B$ are connected to $Y_j$, and then go on to the next basic block. Note that the successful SEQ apparently gives much more information than how many inputs in $B$ should connect to $Y_j$ — it may also suggest connectivity to $Y_j$ for many of the other inputs in the subnetwork. However, it is only for the basic block $B$ that connectivity in the subnetwork implies connectivity in the overall network, since this was how we chose the induced subnetwork.

Thus, the for each basic block and each output, we need at most $k^{2k}\ell^k$ SEQ's, each on a network with at most $k+1$ outputs; since there are $n$ outputs and at most $m$ basic blocks, we obtain:

**Theorem 3.2** *For any class of fan-in $k$, identical input bias, two-layer noisy-OR networks with at most $\ell$ weight values, there is an algorithm for learning a network that is structurally equivalent to an unknown target network from the class with $m$ inputs and $n$ outputs using at most $mnk^{2k}\ell^k$ SEQ's, each on a network with $m$ inputs and $k+1$ outputs.*

The important observations at this point are that the number of queries required is exponential in $k$, but only *polynomial* in $m$ and $n$, despite the fact that the number of networks in the classes considered is exponential in $m$ and $n$; and that the required SEQ's are on small sets of outputs.

## 4  Implementing SEQ's from Data

In order to implement an oracle for SEQ's given only sample data from the outputs of the target network, we will take an obvious approach: given a query on a network with outputs $Y_{j_1}, \ldots, Y_{j_r}$, we will sample the target network distribution restricted to these outputs. If the observed distribution on these outputs differs "significantly" from that defined by the query network, we declare the query incorrect and return NO, and otherwise we declare the query correct and return YES. Thus, in order to implement SEQ's, we will need that (sub)networks that are not structurally equivalent generate different distributions, and furthermore that the difference would be noticeable from a small sample.

For the noisy-OR networks we examine, it turns out to be most convenient to express the distribution on the outputs as a set of polynomials over the uniform input bias $p$; this polynomial is defined by the network structure and weights. As long as networks that are not structurally equivalent give rise to non-identical sets of polynomials, we will be able to argue that we can distinguish different networks on the basis of sample data, and thus implement the desired SEQ's.

### 4.1  Polynomials for Noisy-Or Distributions

Consider a noisy-OR network with $r$ output units $\mathcal{Y} = \{Y_1, \ldots, Y_r\}$ and identical input biases $p$. We would like to compute the probability that the outputs in $\mathcal{Y}$ are all 0 simultaneously. Recall that each input $X_i$ is connected to the outputs in $T_i$. Using the notation of Equation (1), in order for all the outputs in $\mathcal{Y}$ be 0, we need that for any input $X_i$ that is set to 1 and is connected to an output $Y_j \in \mathcal{Y}$, $R_{ij} = 0$. For a given $X_i$, the probability of this event is $\prod_{j \in T_i \cap \mathcal{Y}} \eta_{ij}$. Since each input $X_i$ has probability $p$ of being 0 and $1-p$ of being 1, we have

$$\mathbf{Pr}[Y_1 = 0, \ldots, Y_r = 0] = \prod_{i=1}^{n} \left( p + (1-p) \prod_{j \in T_i \cap \mathcal{Y}} \eta_{i,j} \right) \quad (4)$$

For a fixed noisy-OR network with outputs $\mathcal{Y}$, we will use $\mathcal{Q}_\mathcal{Y}(p)$ to denote the polynomial of Equation (4). Thus, we view the network weights $\eta_{ij}$ as fixed, with the input bias $p$ as the argument.

Let $C$ be a class of (possibly restricted) noisy-OR networks. We say that $C$ has *unique polynomials* if for any $N^1$ and $N^2$ in $C$ that are not structurally equivalent, there is a set $\mathcal{Y}$ of outputs such that $\mathcal{Q}_\mathcal{Y}^1(p)$ and $\mathcal{Q}_\mathcal{Y}^2(p)$ are not identical (that is, there exists a $p$ such that $\mathcal{Q}_\mathcal{Y}^1(p) \neq \mathcal{Q}_\mathcal{Y}^2(p)$), where $\mathcal{Q}_\mathcal{Y}^i(p)$ is the polynomial given by Equation (4) for the network $N^i$.

Note that if two distributions agree exactly on the probability that *any* subset of the outputs is simultaneously 0, then the distributions are in fact identical. Thus, if class $C$ does *not* have unique polynomials, then there are two structurally inequivalent networks in $C$ that generate identical output distributions, and we could never hope to implement SEQ's for this class, or more generally, to exactly learn the structure from observed data. On the other hand, if $C$ does have unique polynomials, we still have work to do, since this only guarantees that for *some* set of the outputs, and for *some* value of the input bias $p$, there is a *non-zero* difference between the probability of all 0's. The first problem — that we don't know which set of outputs yields the differing polynomials — we have essentially already solved, since we have shown how to limit our attention to only $k+1$ outputs at a time in the required SEQ's. The latter two problems — that we only have a guarantee of a difference for some value of $p$, and that this difference may be too small — are tackled in the next section. For now, we simply show several restricted classes of noisy-OR networks with unique polynomials.

We start with the class of noisy-OR networks with just

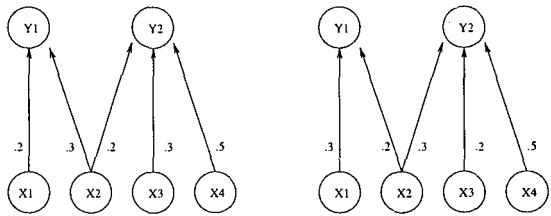

Figure 1: An example showing two noisy-OR networks on inputs $X_1, \ldots, X_4$ and outputs $Y_1, Y_2$ that have only $\ell = 3$ weight values, are not structurally equivalent, yet generate identical distributions on $Y_1$ and $Y_2$ for any input bias $p$.

one weight value.

**Lemma 4.1** *The class of noisy-OR networks with one weight value has unique polynomials.*

This simple example already includes networks of standard logical OR gates, in which the allowed weight value is 0. Unfortunately, it is possible to show that this lemma cannot be generalized to allow even three weight values — that is, there exist two (rather small) noisy-OR networks that are not structurally equivalent, yet generate identical output distributions (see Figure 1). However, in the full paper, we will demonstrate several other restrictions on noisy-OR networks that do yield classes with unique polynomials. One example is the class of networks in which each output $Y_j$ is associated with just a single weight value $\eta_j$ — thus, if $X_i$ is connected to $Y_j$, then $\eta_{ij} = \eta_j$. A similar condition associating each input with just a single weight value also yields unique polynomials. Another sufficient condition is that every weight $\eta_{ij}$ is one of two fixed values $\eta$ or $\xi$, with only $k$ inputs to each output having weight $\eta$. This permits networks in which *every* input influences each output, but with most influences being "weak" (details omitted).

The important points are that the condition of unique polynomials is *required* in order meet the strong learning criterion we are aiming for, and that this condition is met for certain natural restrictions on the networks, but certainly not all. For those classes with unique polynomials, the next section provides rather general methods for implementing SEQ's.

### 4.2 SEQ's from Unique Polynomials

Let us briefly review where we are. We first gave an algorithm assuming SEQ's that learned a network structurally equivalent to the unknown target network, and made SEQ's on subnetworks with only $k + 1$ outputs. We then introduced the notion of a class of networks having unique polynomials, argued that it was required in order to meet our learning criterion from sample data, and demonstrated some classes having unique polynomials. In this section, we show that if a class has unique polynomials, then we can in fact implement SEQ's for "most" values of the input bias $p$.

We first state a general result about polynomials.

**Theorem 4.2** *Let $Q_1(p), \ldots, Q_r(p)$ be univariate polynomials, each with degree at most $d$ and leading coefficient at least $c$. Then for any $\alpha$, the number of distinct values of $p \in [0, 1]$ for which there exists an $i$ satisfying $Q_i(p) = \alpha$ is at most $dr$, and the measure of the set $\{p : \exists i | Q_i(p)| \leq \alpha\}$ is at most $8dr(\alpha/2c)^{1/d}$.*

We omit the proof of this theorem, but it relies on some classical results in approximation theory [R69]. We now show how this result can be used to implement SEQ's from sample data for restricted classes of noisy-OR networks.

Let $C$ be a class of noisy-OR networks with unique polynomials, and let $p \in [0, 1]$. We say that $p$ is $\alpha$-*good* for $C$ if for any two networks $N^1, N^2 \in C$ that are not structurally equivalent, there is a subset $\mathcal{Y}$ of the outputs such that

$$|\mathbf{Pr}_{N^1_p}[\forall Y \in \mathcal{Y} : Y = 0] - \mathbf{Pr}_{N^2_p}[\forall Y \in \mathcal{Y} : Y = 0]| \geq \alpha. \quad (5)$$

In other words, a "good" bias is one that ensures that any two non-equivalent structures have significantly different output distributions.

The following result states that "most" values of the bias $p$ are "reasonably" good for classes of networks with unique polynomials.

**Theorem 4.3** *Let $C$ be a class of noisy-OR networks with fan-in $k$ and $\ell$ weight values multiplying $\beta$. Suppose that $C$ has unique polynomials. Then the measure of $p \in [0, 1]$ that are $\alpha$-good for $C$ is at least $1 - \delta$, for a value of $\alpha$ that is polynomial in $1/k^{k^4}$, $1/\ell^{k^4}$, $1/\beta^{k^3}$, and $1/\delta^{k^2}$.*

**Proof:**(Sketch) Since $C$ has unique polynomials, for any networks $N^i$ and $N^j$ in $C$ we know that there exists a set $\mathcal{Y}$ of their outputs such that the polynomials $\mathcal{Q}_\mathcal{Y}^i(p)$ and $\mathcal{Q}_\mathcal{Y}^j(p)$ are not identical. Furthermore, by the basic block arguments already given, we may assume that $|\mathcal{Y}| \leq k + 1$. This implies that $\mathcal{R}_\mathcal{Y}^{i,j}(p) = \mathcal{Q}_\mathcal{Y}^i(p) - \mathcal{Q}_\mathcal{Y}^j(p) \not\equiv 0$. If in addition, for a given value of $p$, $|\mathcal{R}_\mathcal{Y}^{i,j}(p)| > \alpha$, this $p$ is good for this $N^i$ and $N^j$; if this holds for any $N^i$ and $N^j$ in the class $C$, we conclude that this $p$ is $\alpha$-good for $C$. The proof proceeds by applying Theorem 4.2 to the set of all $\mathcal{R}_\mathcal{Y}^{i,j}(p)$, with bound on the number of such polynomials derived from the restrictions on the weight values.
□

Thus, provided $C$ has unique polynomials, we can implement SEQ's from observed output data for "most" (but not all) values of the input bias. Provided $p$ is $\alpha$-good, we can answer (with high probability) any SEQ by simply sampling sufficiently to determine if there is some subset of the outputs on which the distribution defined by the queried subnetwork differs from the target distribution by more than $\alpha$.

### 4.3 Wrapping Things Up

The combination of results from the previous sections finally yields the following.

- For the restricted classes of noisy-OR networks discussed in Section 4.1, we have algorithms that will, with probability $1 - \delta$, derive from a sufficiently large random sample of output values a network that is structurally equivalent to the target network, for "most" values of the input bias.

- The algorithms all reconstruct the target network incrementally, always deciding how to add a new output through a series of SEQ's defined by the current basic blocks. (Section 3.2)

- The SEQ's are implemented by sampling sufficiently from the target output distribution to see if the queried subnetwork generates a distribution sufficiently similar to assert structural equivalence. (Section 4; Theorem 4.3)

- The probability $1 - \delta$ of success by the algorithms is taken over the draw of the random sample. The running time and sample size required by the algorithms will depend only polynomially on the number of outputs of the target network, but exponentially on the fan-in.

- The measure of the set of input biases for which the algorithms succeed can be made arbitrarily close to 1 at the expense of increased running time. (Theorem 4.3)

More formally, we can establish the following theorem. Similar results can be stated for the other classes of noisy-OR networks that have unique polynomials, discussed in Section 4.1.

**Theorem 4.4** *Let $C$ be the class of noisy-OR networks with one weight value $\eta$ and fan-in $k$. There exists an algorithm $A$, such that for any $N \in C$, and for the input bias $p$ chosen at uniformly in $[0, 1]$, the algorithm $A$, given $p$ and $\eta$ as input, and given access to random examples generated according to the output distribution $N_p$, produces a network $N'$, such that with probability $1 - \delta$ the networks $N$ and $N'$ are structurally equivalent. (Here the probability is both over the choice of $p$ and the random sample.) Furthermore, the running time of $A$ is polynomial in $m$ (the number of inputs of $N$), $n$ (the number of outputs of $N$), $1/\delta$ (the confidence), $1/((1 - \eta)\eta)$ (the weight value), and $1/p$ (the input bias), and exponential in $k$ (the fan-in bound). (More precisely, the running time is bounded by $mn(1/p\eta)^{k^c}$, for some constant $c$.)*

## Acknowledgements

Y. Mansour was supported in part by a grant from the Israel Science Foundation.